\documentclass[conference]{IEEEtran}
\IEEEoverridecommandlockouts
\usepackage{cite}
\usepackage{amsmath,amssymb,amsfonts}
\usepackage{algorithm}
\usepackage{algorithmic}

\usepackage{booktabs}
\usepackage{graphicx}
\usepackage{subcaption}
\usepackage{hyperref}  
\usepackage{graphicx}
\usepackage{textcomp}
\usepackage{xcolor}
\def\BibTeX{{\rm B\kern-.05em{\sc i\kern-.025em b}\kern-.08em
    T\kern-.1667em\lower.7ex\hbox{E}\kern-.125emX}}
\begin{document}

\title{
ExARNN: An Environment-Driven Adaptive RNN for Learning Non-Stationary Power Dynamics
\vspace{-3mm}
}

\author{\IEEEauthorblockN{Haoran Li, Muhao Guo, Yang Weng}
\IEEEauthorblockA{\textit{Department of Electrical, Computer and Energy Engineering} \\
\textit{Arizona State University},
Tempe, USA \\
\mbox{\{lhaoran, mguo26, yang.weng\}@asu.edu}}
\and
\IEEEauthorblockN{Marija Ilic, Guangchun Ruan}
\IEEEauthorblockA{\textit{Department of Electrical, Computer and Energy Engineering} \\
\textit{Massachusetts Institute of Technology},
Cambridge, USA \\
\mbox{\{ilic, gruan\}@mit.edu}}
}

\maketitle

\begin{abstract}

Non-stationary power system dynamics, influenced by renewable energy variability, evolving demand patterns, and climate change, are becoming increasingly complex. Accurately capturing these dynamics requires a model capable of adapting to environmental factors. Traditional models, including Recurrent Neural Networks (RNNs), lack efficient mechanisms to encode external factors, such as time or environmental data, for dynamic adaptation. To address this, we propose the External Adaptive RNN (ExARNN), a novel framework that integrates external data (e.g., weather, time) to continuously adjust the parameters of a base RNN. ExARNN achieves this through a hierarchical hypernetwork design, using Neural Controlled Differential Equations (NCDE) to process external data and generate RNN parameters adaptively. This approach enables ExARNN to handle inconsistent timestamps between power and external measurements, ensuring continuous adaptation. Extensive forecasting tests demonstrate ExARNN's superiority over established baseline models.   
\end{abstract}


\section{Introduction}
With the increasing integration of renewable energy and variable demands, power systems have growing non-stationary dynamics affected by the external environment \cite{10521705}. For example, renewable sources like wind and solar are inherently variable, as their generation capacity depends on fluctuating weather conditions, such as sunlight intensity and wind speed \cite{ahmed2019review,guo2023graph}. Similarly, electricity demand is influenced by factors like temperature and time of day \cite{7529083}, leading to significant variations in load patterns. 

These external dependencies make traditional models, which assume stationarity \cite{ahmad2022load}, increasingly inadequate for capturing the real-time dynamics of power systems. As a result, information fusion or adaptive modeling approaches that can account for non-stationary behaviors are critical for reliable forecasting, control, and optimization in modern power grids. In particular, in the era of big data, the emerging system state and environmental measurements \cite{safdarian2024detailed,10636963,10415845} make it possible to create fused or adaptive models. 

Classical methods for load and generation forecasting have traditionally integrated weather data to capture linear dependencies between meteorological conditions and power demand or renewable generation. ARIMAX models \cite{hyndman2018forecasting}, for instance, extend the ARIMA framework by including exogenous variables like temperature, humidity, and wind speed, capturing straightforward correlations between weather and system loads. Similarly, linear regression models incorporate weather features, often selected based on historical correlations, to model demand fluctuations. Kalman filters and state-space models are also used, where weather data contributes to real-time state estimation of system dynamics \cite{al2018short}. However, these classical methods are limited by their assumptions of linearity and stationarity, which restrict their adaptability to the non-linear and dynamic relationships seen in modern power systems with high renewable penetration. While efficient and interpretable, these models often require significant manual feature engineering to improve performance, particularly in non-stationary environments where weather patterns introduce complex variations over time \cite{10163982,fan2011short,cui2023sig2vec,guo2023msq,guo2024bayesian}.

Machine Learning (ML) and Deep Learning (DL) approaches have greatly advanced forecasting by capturing non-linear interactions and temporal dependencies, making them well-suited for weather data fusion. ML techniques like Support Vector Regression (SVR) and ensemble methods \cite{zendehboudi2018application} accommodate more complex, non-linear relationships between weather and power demand or generation but often rely on manually crafted features to account for sequential dependencies. Deep learning models, such as Recurrent Neural Networks (RNNs) and Long Short-Term Memory (LSTM) networks, enhance forecasting accuracy by learning from historical load and weather data in a sequential manner, capturing dependencies across time \cite{kong2017short,9996971}. Hybrid models combining CNNs with RNNs also integrate spatial weather data, such as satellite images, for renewable generation forecasting \cite{shi2015convolutional}.

However, sample efficiency is a key issue for the above methods, given that power system and environmental data have inconsistent measurement quality and density. Power data, including load and generation metrics, is typically collected at high frequencies, providing dense and consistent information streams \cite{weron2014electricity,10138375}. In contrast, weather data, especially localized meteorological observations like solar irradiance or wind speeds, often lacks the temporal resolution needed to match power system data \cite{bessa2019handling}. This discrepancy complicates the data fusion process, as integrating high-density power data with lower-frequency weather inputs can lead to alignment issues and gaps in capturing dynamic dependencies.

In this paper, we propose a novel DL model, dubbed ExARNN, to continuously adapt a base RNN model by efficiently leveraging environmental data. The goal is achieved by the following key designs. $(1)$ We treat external information as meta-knowledge to adjust a base RNN, following the idea of hypernetwork \cite{ha2016hypernetworks}. This reduces the risk of overfitting to specific environmental conditions. In contrast, directly inputting weather data would make the model perform poorly when encountering new or slightly different weather patterns. $(2)$ We employ a continuous process, Neural Controlled Differential Equation (NCDE) \cite{kidger2020neural},  to fully integrate low-frequency weather data, maximizing the sample efficiency.   

In general, we have the following contributions. $(1)$ We formalize a learning problem to integrate power system and external environment measurements to train a DL model for approximating power system non-stationary dynamics. $(2)$ We solve the problem by proposing ExARNN, a highly generalizable and sample-efficient DL model. $(3)$ We conduct experiments to validate the high performance of ExARNN.

\section{Problem Formulation}

In this section, we formalize the problem as follows.
\begin{itemize}
\item Problem: Building an adaptive DL model for non-stationary power system dynamics. 
\item Given: Power system measurements $\{\boldsymbol{x}(t_i))\}_{i\in\mathcal{N}_x}$ and environmental measurements $\{\boldsymbol{w}(t_i)\}_{i\in\mathcal{N}_w}$, where in most scenarios, $\mathcal{N}_w\subseteq \mathcal{N}_x$. 
\item Find: A well-trained model $f_{\theta(\boldsymbol{w})}(\cdot)$ to utilize historical data to predict $\hat{\boldsymbol{x}}(t_{i+1})$, where $\theta(\boldsymbol{w})$ is a function of $\boldsymbol{w}$ to achieve parameter adaptation according to external data. 
\end{itemize}

To achieve this goal, we demand a mechanism to understand the context information in $\boldsymbol{w}$ and generate $\theta(\boldsymbol{w})$ for the main RNN. Moreover, all the inconsistent data should be efficiently used to maximize the sample efficiency. 

\vspace{-2mm}

\section{Preliminary}

\subsection{Hypernetwork and Its Application to RNN}
Hypernetwork \cite{ha2016hypernetworks}  is a neural network architecture where one network (the hypernetwork) generates the weights for another network. Let $\boldsymbol{w}$ be the input to the hypernetwork, which represents the external factor. Let $g(\cdot)$ denote the hypernetwork that generates (partial) parameters of the main network. We have $\theta_1(\boldsymbol{w})=g(\boldsymbol{w})$, where $\theta_1\subseteq \theta$. To incorporate flexibility, we treat $\theta_0 = \theta \setminus \theta_1$ as static parameters independent of environments. 

For an RNN model $f_{\theta(\boldsymbol{w})}(\cdot)$, applying hypernetwork has been well investigated in \cite{ha2016hypernetworks}. Let $\boldsymbol{h}_i$ denote the hidden state of the RNN $f_{\theta(\boldsymbol{w})}(\cdot)$ in time $t_i$. We formulate the hypernetwork-controlled RNN as follows:
\begin{equation}
\begin{aligned}
\label{eqn:hyp_rnn}
\boldsymbol{h}_i&=\sigma_1\big(W_{1}\boldsymbol{x}(t_i)+\theta_1(\boldsymbol{w})\boldsymbol{h}_{i-1}+b_1\big),\\
\hat{\boldsymbol{x}}(t_{i+1}) &= \sigma_2(W_{2}\boldsymbol{h}_i+b_2),
\end{aligned}
\end{equation}
where $\sigma_1$ and $\sigma_2$ are activation functions, $W_1$, $W_2$, $b_1$, and $b_2$ are static weight and bias terms, i.e., $\theta_0=\{W_1, W_2, b_1, b_2\}$. In this formulation, \textbf{$\theta_1(\boldsymbol{w})$, as the output of $g(\boldsymbol{w})$, dynamically determines how much information from past hidden state $\boldsymbol{h}_{i-1}$ can be passed to $\boldsymbol{h}_i$, which is subsequently utilized to predict $\hat{\boldsymbol{x}}(t_{i+1})$.} To summarize, the environmental data provides meta-context information to drive the RNN to understand the past information to predict the future. 

\noindent \textbf{Remark}: The base RNN model can be replaced with other sequence models like LSTM. However, we find that their performance in load forecasting is very similar.

The choice for the hypernetwork $g(\boldsymbol{w})$ becomes another issue, especially when the measurements $\{\boldsymbol{w}(t_i)\}_{i\in\mathcal{N}_w}$ are much more sparse compared to $\{\boldsymbol{x}(t_i))\}_{i\in\mathcal{N}_x}$. Thus, one can hardly obtain $\theta_1(\boldsymbol{w}(t_j))$, where $i=\mathcal{N}_x\setminus\mathcal{N}_w$ to construct the main RNN. Consequently, we propose to leverage NCDE to create a continuous feature flow that can be mapped to $\theta_1(t_j)$ at arbitrary time $t_j$.

\subsection{Neural Controlled Differential Equations}
\label{subsec:ncde}
NCDE is an extension of Neural ODE (NODE) \cite{chen2018neural} to process sequential data. Therefore, we first introduce NODE. NODE is a continuous version of the ResNet \cite{he2016deep}. In ResNet, a neural network building block is used to approximate the difference between two layer's hidden features. Correspondingly, NODE approximates the derivative of the input hidden feature. Specifically, we can formulate NODE as follows:
\begin{equation}
\label{eqn:NODE}
\begin{aligned}
\boldsymbol{z}(t_1) &= l_1\big(\boldsymbol{w}(t_1)\big),\\
\boldsymbol{z}(t_2)&=\text{Solve}\big(\boldsymbol{z}(t_1),h,t_2\big)=\boldsymbol{z}(t_1) + \int_{t_1}^{t_2}h\big(\boldsymbol{z}(t)\big)dt,\\
\theta_1(t_2)&=l_2\big(\boldsymbol{z}(t_2)\big),
\end{aligned}
\end{equation}
where $l_1(\cdot)$ and $l_2(\cdot)$ are two fully-connected layers, $\boldsymbol{z}$ represents a vector of hidden features, $h(\cdot)$ is a neural network to map $\boldsymbol{z}$ to the approximation of $\dot{\boldsymbol{z}}$, and $\text{Solve}(\cdot)$ is the operation to employ an ODE solver to solve the Initial Value Problem (IVP). Then, $\boldsymbol{z}(t_1)$ as the output feature can be mapped to $\theta(\boldsymbol{w})$. Nevertheless, the mapping, although operating on a continuous feature flow $\boldsymbol{z}(t)$ in Equation \eqref{eqn:NODE}, doesn't efficiently incorporate the existing environmental information because $\boldsymbol{z}(t)$ only depends on the initial value. 

To fix the issue, NCDE demonstrates how to integrate the continuous information based on the rough path theory \cite{lyons2007differential}. Specifically, NCDE first creates a natural cubic spline such that $W(t_i)=(\boldsymbol{w}(t_i),t_i)$. In rough path theory, $W(t)$ is a continuous path that can continuously control the flow of $\boldsymbol{z}(t)$ with the so-called Riemann–Stieltjes integral:

\begin{equation}
\label{eqn:NCDE}
\begin{aligned}
\boldsymbol{z}(t_2,W)=\boldsymbol{z}(t_1) + \int_{t_1}^{t_2}\hat{h}\big(\boldsymbol{z}(s)\big)\frac{dW(s)}{ds}(s)ds,
\end{aligned}
\end{equation}
where $\hat{h}$ is a neural network that outputs a matrix to describe a vector field. The derivative $\frac{dW(s)}{ds}(s)$ works as a control signal to continuously drive the dynamics of $\boldsymbol{z}(t)$. Hence, we utilize Equation \eqref{eqn:NCDE} to replace the feature evolution in Equation \eqref{eqn:NODE} with continuous control from the cubic spline $W(t)$. More importantly, $\boldsymbol{z}(t_2,W)$ can be evaluated at arbitrary time $t_2$ and output $\theta_1(t_2,W)$ for the main RNN model.

\section{Proposed Model}
\label{sec:model}
{\color{blue}We demonstrate a powerful ExARNN model to (1) process inconsistent resolutions and (2) achieve adaptations. The former target is achieved by employing an ODE solver to create a continuous ODE flow to depict the environment feature dynamics, where the derivative of the feature is approximated by a neural network. Then, by computing the integration using the ODE solver and the neural network, the environment feature can be evaluated at \emph{arbitrary times}. The latter target is gained by developing \emph{an adaptive learning model that can adjust itself according to time and external weather data}.}
\subsection{Architecture of ExARNN}\label{subsec:archi}

\begin{figure}
    \centering    \includegraphics[width=0.9\linewidth,trim={270 100 250 60}]{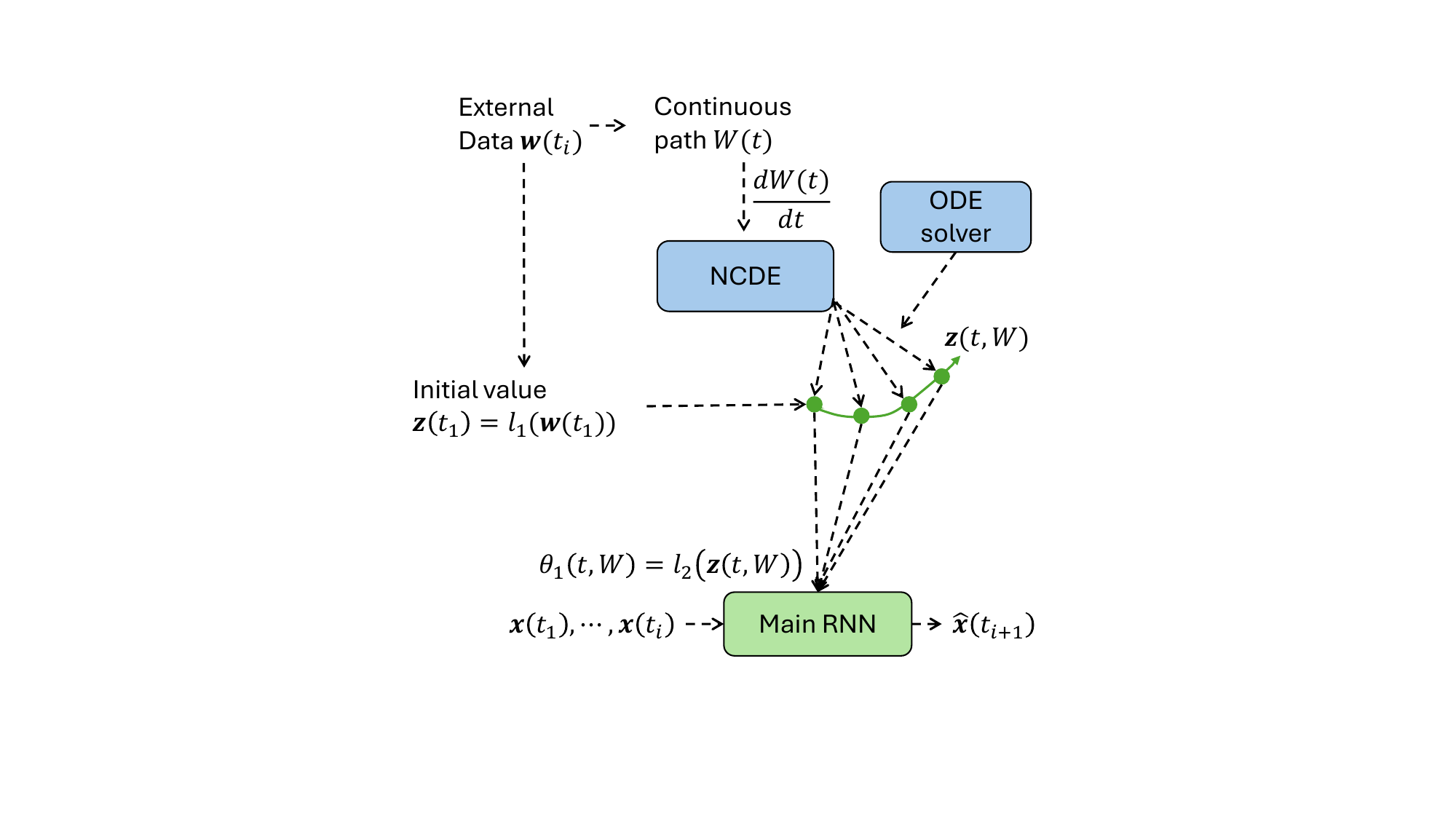}
    \caption{The main framework of the proposed ExARNN.}
    \label{fig:main}
    \vspace{-6mm}
\end{figure}

Fig. \ref{fig:main} demonstrates the architecture of ExARNN that automatically adapts the RNN model based on external data. NCDE model leverages the ODE solver (blue blocks) to process the continuous path $W(t)$ for the input environmental data and control the evolution of the flow $\boldsymbol{z}(t,W)$ (solid green arrows with green points that represent feature values evaluated at time $t_i$, where $i\in\mathcal{N}_x$). Subsequently, $l_2(\cdot)$ map $\boldsymbol{z}(t_i,W)$ to $\theta_1(t_i,W)$ for arbitrary $i\in\mathcal{N}_x$ which, combined with static parameter set $\theta_0$, can together shape the parameter space of the main RNN (green block). Below are the steps.

\begin{itemize}
    \item Step 1. Create continuous process $W(t)$. With data $\{\boldsymbol{w}(t_i)\}_{i\in\mathcal{N}_w}$, we employ a natural cubic spline to generate $W(t)$. 
    \item Step 2. Compute the initial value $\boldsymbol{z}(t_1)=l_1(\boldsymbol{w}(t_1))$. The initial value of the feature state is mapped from the initial value of the environmental measurements with a fully-connected layer $l_1(\cdot)$.
    \item Step 3. Employ an ODE solver to create the feature flow $\boldsymbol{z}(t,W)$. By Equation \eqref{eqn:NCDE}, we utilize an ODE solver to compute the integral and create the feature flow that can be evaluated at times $\{t_i\}_{i\in\mathcal{N}_x}$. 
    \item Step 4. Construct the main RNN at $\{t_i\}_{i\in\mathcal{N}_x}$. Specifically, we compute $\theta_1(t_i,W)=l_2(\boldsymbol{z}(t_i,W))$ and use $\theta_1(t_i,W)$ and $\theta_0$ to build the main RNN, shown in Equation \eqref{eqn:hyp_rnn}. 
\end{itemize}

We package the above process as a subprogram called Build-ExARNN. 
{\color{blue}To achieve multi-resolution data processing, as shown in Fig. 1, we utilize an NCDE to input weather data and output the weather feature flow $\boldsymbol{z}(t,W)$ in the green curve. This is achieved by using an ODE solver to compute the integration of a derivative function that is approximated by a neural network, illustrated in Equations \eqref{eqn:NODE} and \eqref{eqn:NCDE} and Steps 1-3. Then,  we utilize again the ODE solver to compute the integral and evaluate the environment features at time $\{t_i\}_{i\in\mathcal{N}_x}$, i.e., all timestamps at the power system measurements. Hence, the environment feature and the power system measurements are synchronized. Subsequently, we utilize another layer $l_2$ to map from weather features to a parameter set for the main RNN to construct the main RNN at $\{t_i\}_{i\in\mathcal{N}_x}$. For example, if we construct the main RNN at time $t_i$, this RNN will be used to predict the power system data $\hat{\boldsymbol{x}}(t_{i+1})$. The mathematical formulations are depicted in Equation \eqref{eqn:hyp_rnn} and Step 4. Eventually, the main RNN has a parameter set $\theta_1(t,W)$ that depends on the weather and time, gaining the adaptation capacity.}

\subsection{Training Algorithm for ExARNN}

After the subprogram Build-ExARNN, we have the ExARNN model that outputs $\hat{\boldsymbol{x}}(t_{i+1})$ to approximate $\boldsymbol{x}(t_{i+1})$. Thus, the training loss is the Mean Square Error (MSE):
\begin{equation}
\label{eqn:train_mse}
L_{MSE}=\sum_{i\in\mathcal{N}_x\setminus{\{1\}}}||\hat{\boldsymbol{x}}(t_i)-\boldsymbol{x}(t_i)||_2^2,
\end{equation}
where $||\cdot||_2$ is the $L^2$ norm. In ExARNN with a hypernetwork architecture, we need to decide what parameters are updated during training. Let $\psi$ denote the set of parameters for $l_1(\cdot)$, $l_2(\cdot)$, and $\hat{h}(\cdot)$ in Equations \eqref{eqn:NODE} and \eqref{eqn:NCDE} for the NCDE model. $\psi$ should be updated during training to generate better $\theta_1$ for the main RNN. To maintain consistency during training, $\theta_1$ in the main RNN shouldn't be updated because it's only determined by NCDE. Consequently, the whole trainable parameter is $\{\psi, \theta_0\}$, where $\theta_0$ as the static parameters is defined in Equation \eqref{eqn:hyp_rnn}. In general, we propose Algorithm \ref{alg:train_ExARNN}.

\begin{algorithm}
\caption{Training algorithm for ExARNN.}
\label{alg:train_ExARNN}

\textbf{Function:} $\text{Train-ExARNN}\big(\{\boldsymbol{x}(t_i))\}_{i\in\mathcal{N}_x}, \{\boldsymbol{w}(t_i)\}_{i\in\mathcal{N}_w}\big)$.

\textbf{Input:} Measurements $\{\boldsymbol{x}(t_i))\}_{i\in\mathcal{N}_x}$ and $\{\boldsymbol{w}(t_i)\}_{i\in\mathcal{N}_w}$.

\textbf{Hyper-parameters:} Learning rate $\eta$ and number of epoch $N$.

\begin{algorithmic}[1]
\STATE \textbf{Model construction}. Use Build-ExARNN in Section \ref{subsec:archi} to construct the ExARNN model.

\FOR{$n\leq N$}
\STATE \textbf{Compute loss}. Use Equation \eqref{eqn:train_mse} to compute the loss.
\STATE \textbf{Parameter updates with the rule}:
\begin{equation}
\begin{aligned}
\psi^{(n+1)} &\leftarrow \psi^{(n)}- \eta \frac{\partial L_{MSE}}{\partial \psi}|_{\psi=\psi^{(n)}},\\
\theta_0^{(n+1)} &\leftarrow \theta_0^{(n)} - \eta \frac{\partial L_{MSE}}{\partial \theta_0}|_{\theta_0=\theta_0^{(n)}}.\\
\end{aligned}
\end{equation}
\ENDFOR
\end{algorithmic}
\textbf{Output:} The well-trained ExARNN.
\end{algorithm}

\section{Experiment}
\subsection{Settings}
\begin{figure}
    \centering \includegraphics[width=1.0\linewidth]{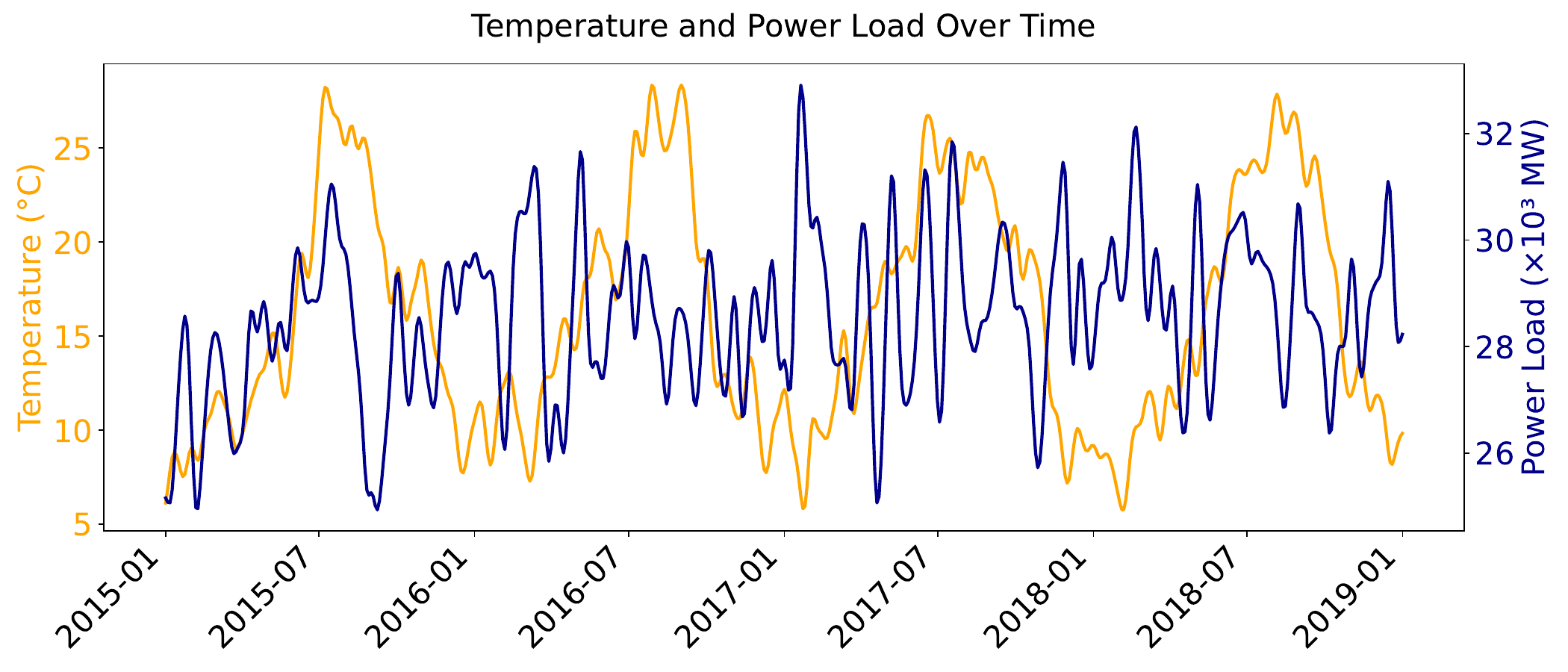}
    \caption{Data of load and temperature in Spain.}
    \label{fig:raw_load_temp_spanish}
    \vspace{-4mm}
\end{figure}

We evaluate our model using two power system datasets. The Spain dataset \cite{rolnick2022tackling} comprises four years of data on electrical consumption, generation, pricing, and weather conditions for Spain. The consumption and generation data were obtained from ENTSOE, a public platform for Transmission System Operator (TSO) data. Weather information was sourced from the Open Weather API for the five largest cities in Spain as part of a personal project and made publicly available. Fig. \ref{fig:raw_load_temp_spanish} visualizes the load and temperature data for Spain, which suggests strong correlations. The Texas dataset includes load data provided by one of our partner utilities in Texas in the U.S. Corresponding temperature data for Texas, aligned with the same timestamps, was collected using the Visual Crossing Weather API History Data. \textbf{The resolution for weather and power data is $60$ and $15$ minute/sample, respectively}.


To demonstrate the high performance of ExARNN, we introduce the following benchmark methods. (1) Vanilla RNN model. The classic RNN model can input power system/environmental data and output the predicted system data. To guarantee that all data is used, we utilize a cubic spline to interpolate low-frequency environmental data to align the sample resolution. (2) RNN-$\Delta_t$. To fully use low-frequency environmental and high-frequency power data, one can embed a continuous ODE process, where all data can be used with the direct encoding of the time difference, i.e., $\Delta_t$ \cite{che2018recurrent}. (3) ODE-RNN \cite{rubanova2019latent}. Neural ODE is embedded to learn the continuous dynamical process of hidden states. (4) NCDE \cite{kidger2020neural}. As discussed in Section \ref{subsec:ncde}, NCDE can create a continuous process and fully use all the data.

\vspace{-3mm}
\subsection{General Results}
\begin{figure*}
    \centering
    \begin{subfigure}{\linewidth}
        \centering
        \includegraphics[width=0.94\linewidth, trim={10 10 10 1}]{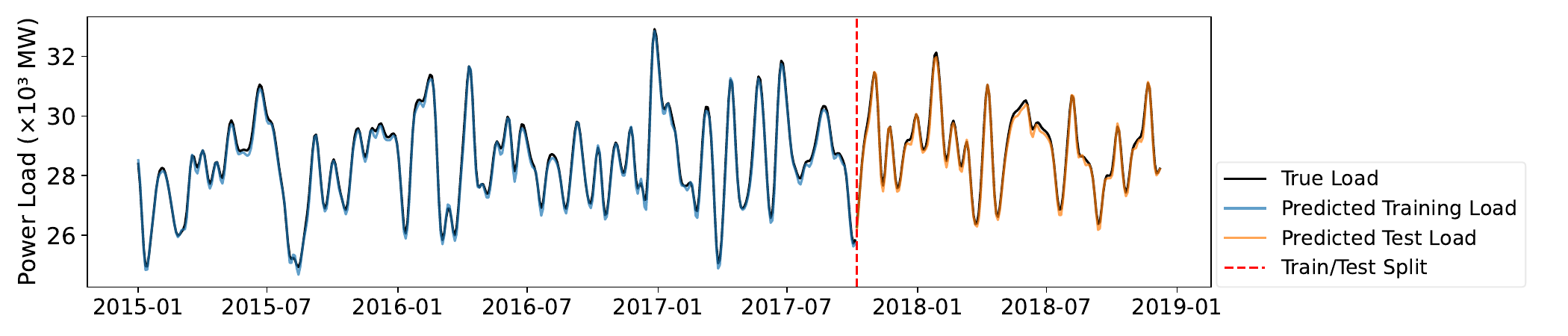}
        \caption{The result of ExARNN.}
        \label{fig:subfig5}
    \end{subfigure}
    
    \begin{subfigure}{0.49\linewidth}
        \centering
        \includegraphics[width=\linewidth, trim={10 10 10 1}]{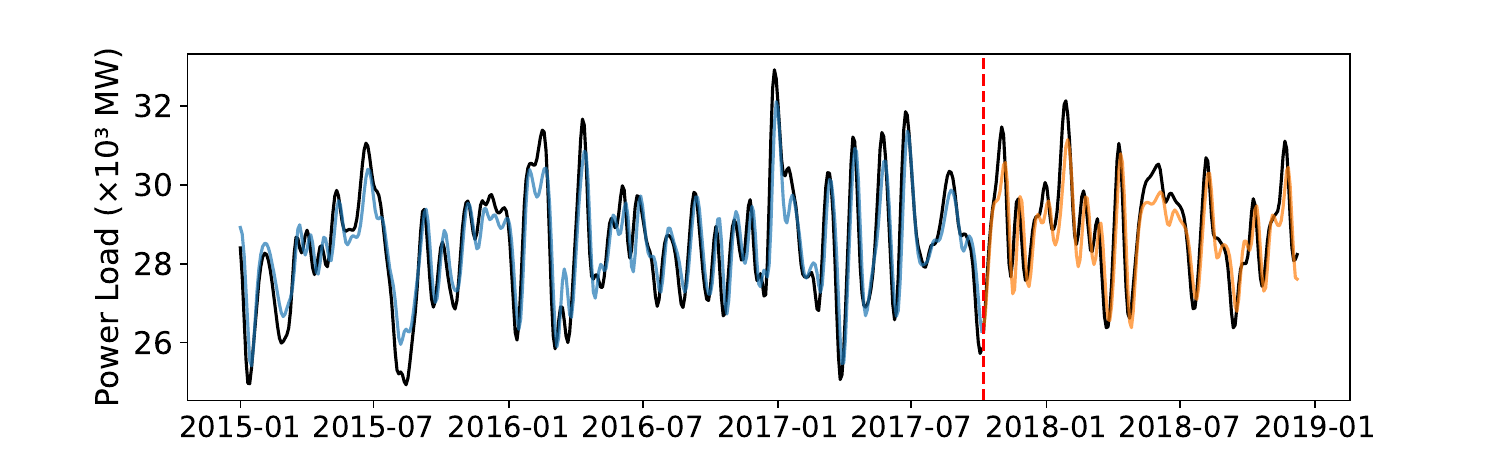}
        \caption{The result of RNN.}
        \label{fig:subfig1}
    \end{subfigure}
    \begin{subfigure}{0.49\linewidth}
        \centering
        \includegraphics[width=\linewidth, trim={10 10 10 1}]{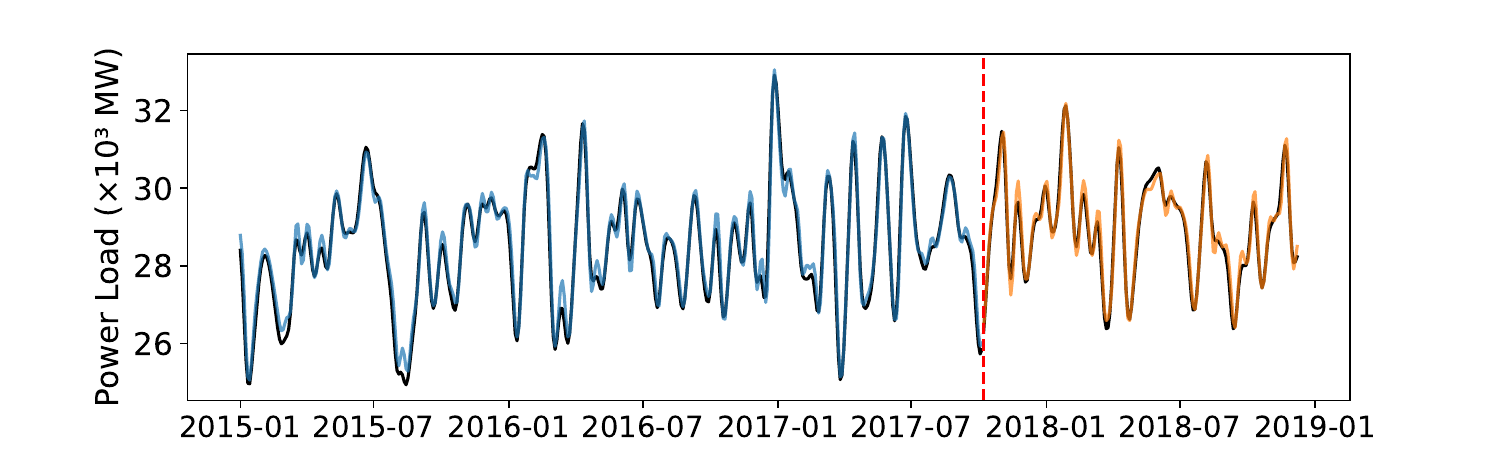}
        \caption{The result of RNN-$\Delta_t$.}
        \label{fig:subfig2}
    \end{subfigure}

    \begin{subfigure}{0.49\linewidth}
        \centering
        \includegraphics[width=\linewidth, trim={10 10 10 1}]{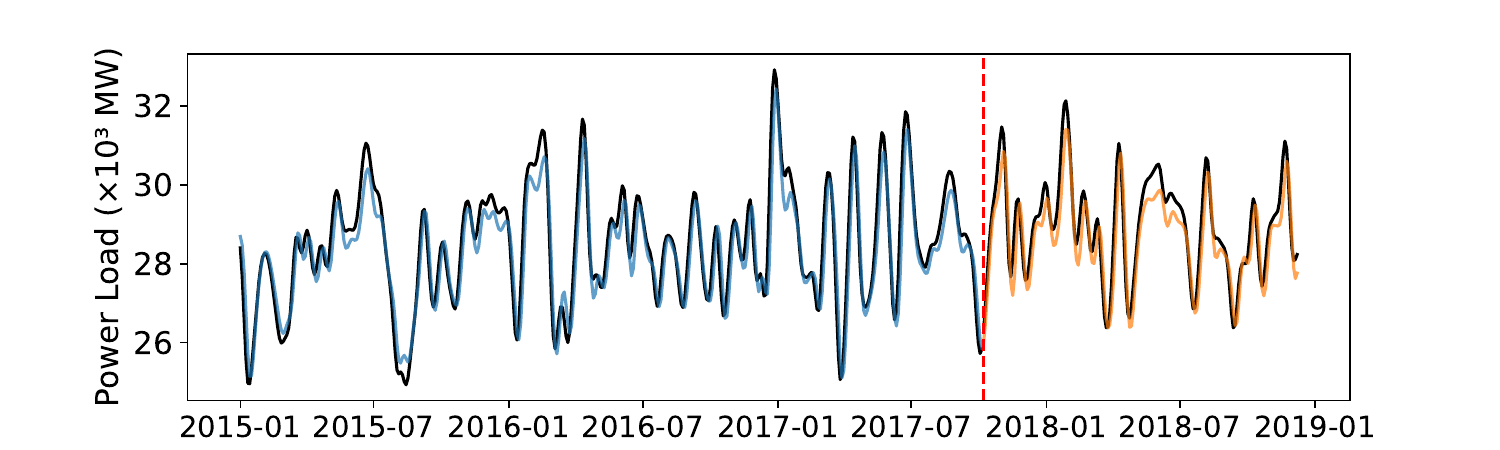}
        \caption{The result of ODE-RNN.}
        \label{fig:subfig3}
    \end{subfigure}
    \begin{subfigure}{0.49\linewidth}
        \centering
        \includegraphics[width=\linewidth, trim={10 10 10 1}]{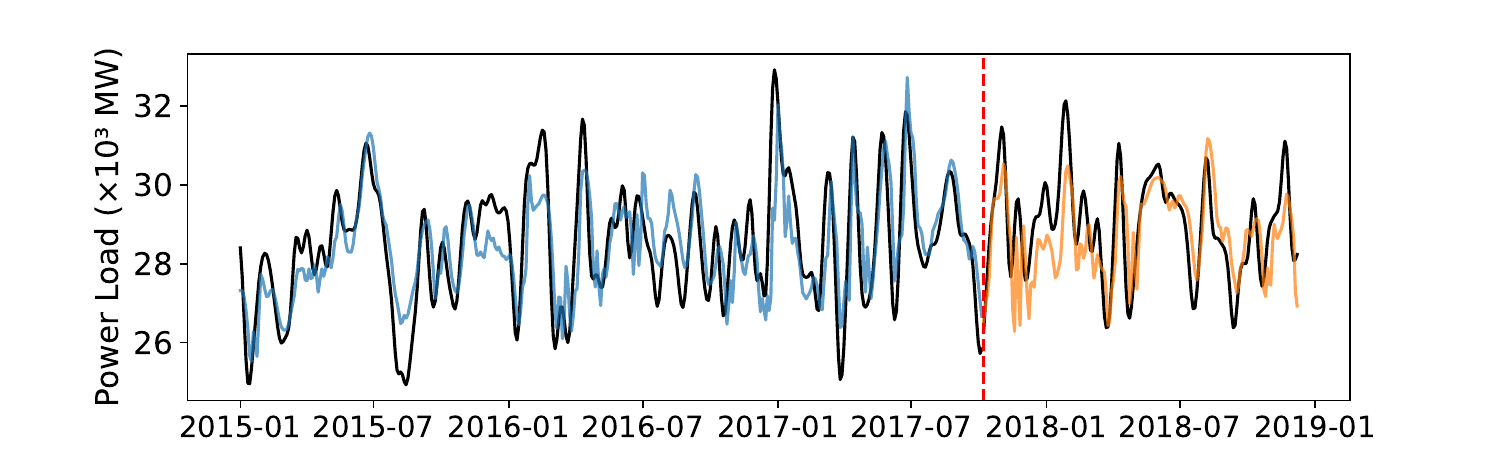}
        \caption{The result of NCDE.}
        \label{fig:subfig4}
    \end{subfigure}

    \caption{Train and test performances of Spain Load Dataset for different methods.}
    \label{fig:overall_figure_spanish}
    \vspace{-4mm}
\end{figure*}

The results, summarized in Table \ref{tab:model_comparison}, show that ExARNN consistently outperformed all other models in both datasets, achieving the lowest MAPE and MSE. Specifically, ExARNN demonstrated superior performance with a MAPE of 1.82\% and an MSE of 0.0001 on the Spanish dataset and a MAPE of 4.64\% and an MSE of 0.0458 on the Texas dataset. Visual comparisons of test performances on the Spanish dataset in Fig. \ref{fig:overall_figure_spanish} further confirm that ExARNN closely aligns with the actual load data, demonstrating more precise and stable predictive capabilities. This highlights ExARNN's potential as a robust and accurate model for power system forecasting.

\begin{table}[h!]
    \centering
    \caption{Prediction error comparison in test data.}
    \begin{tabular}{lcccc}
        \toprule
        \textbf{Model} & \multicolumn{2}{c}{\textbf{Texas}} & \multicolumn{2}{c}{\textbf{Spain}} \\
        \cmidrule(lr){2-3} \cmidrule(lr){4-5}
        & \textbf{MAPE (\%)} & \textbf{MSE} & \textbf{MAPE (\%)} & \textbf{MSE} \\
        \midrule
        RNN & 5.89\% & 0.0801 & 13.74\%  & 0.0064 \\
        RNN-$\Delta_t$ & 4.82\% & 0.0573 & 4.95\% & 0.0008\\
        ODE-RNN & 4.75\% &  0.0575 & 8.97\% &0.0025 \\
        NCDE & 17.78\% & 0.4716 &   19.87\% & 0.0125\\
        \textbf{ExARNN} &  \textbf{4.64\%} &  \textbf{0.0458} &  \textbf{1.82\%} &  \textbf{0.0001}\\
        \bottomrule
    \end{tabular}
    \label{tab:model_comparison}
    \vspace{-5mm}
\end{table}

    




\subsection{Computational Time}

Table \ref{tab:computational_time} compares the test times of various models on the Texas and Spanish datasets. RNN model achieves the fastest inference, followed by RNN-$\Delta_t$ and NCDE. The proposed ExARNN model shows longer times of 0.0383 seconds for Texas and 0.0898 seconds for Spain. Despite not being the fastest, ExARNN's computational cost is balanced by its superior predictive performance, and the time is affordable for power system operations.

\begin{table}[h!]
    \centering
    \caption{Test time comparisons.}
    \begin{tabular}{l|c|c}
        \toprule
        \textbf{Model} & \textbf{Time (s) - Texas} & \textbf{Time (s) - Spain} \\
        \midrule
        RNN &0.0005 & 0.0028\\
        RNN-$\Delta t$ &0.0014& 0.0098\\
        ODE-RNN &0.0359 & 0.0842\\
        NCDE & 0.0046& 0.0133\\
        ExARNN & 0.0383 & 0.0898\\
        \bottomrule
    \end{tabular}
    \label{tab:computational_time}
    \vspace{-5mm}

\end{table}

\section{Conclusion}
In this paper, we propose a novel DL model, ExARNN, to efficiently utilize environmental and power system data to learn non-stationary power system dynamics. Our proposed model is highly sample-efficient and generalizable because we $(1)$ cast the environmental data as meta-knowledge to guide the continuous evolution of the main RNN model and $(2)$ employ NCDE to completely integrate low-frequency environmental data and high-frequency power data in a continuous process. Next, we propose a simple end-to-end training algorithm. The model is compared with many advanced baselines and provides high performances for different tasks. In the future, we will extend the model to a probabilistic setting and deliver more comprehensive evaluations for different dynamic learning tasks in power systems. 

\bibliographystyle{IEEEtran}
\bibliography{IEEEabrv,reference}

\end{document}